\documentclass[12pt]{iopart}
\usepackage{graphicx}
\usepackage{url}
\usepackage{cite}
\newcommand{\figref}[1]{Fig.\ref{#1}}

\newcommand{\eqref}[1]{(\ref{#1})}

\begin{document}

\title{Tensor tree learns hidden relational structures in data to construct generative models}
\author{Kenji Harada\footnote{corresponding author}$^1$, Tsuyoshi Okubo$^2$, and Naoki Kawashima$^{3,4}$}
\ead{harada.kenji.8e@kyoto-u.ac.jp}
\address{
    $^1$ Graduate School of Informatics, Kyoto University, Kyoto 606-8501, Japan
}

\ead{t-okubo@phys.s.u-tokyo.ac.jp}
\address{
    $^2$ Institute for Physics of Intelligence, The University of Tokyo, Tokyo 113-0033, Japan
}

\ead{kawashima@issp.u-tokyo.ac.jp}
\address{
    $^3$ Institute for Solid State Physics, The University of Tokyo, Kashiwa, 277-8581, Japan}
\address{
    $^4$ Trans-scale Quantum Science Institute, The University of Tokyo, Bunkyo-ku, Tokyo 113-0033, Japan
}
\date{April 1, 2025}

\begin{abstract}
    Based on the tensor tree network with the Born machine framework, we propose a general method for constructing a generative model by expressing the target distribution function as the amplitude of the quantum wave function represented by a tensor tree. The key idea is dynamically optimizing the tree structure that minimizes the bond mutual information. The proposed method offers enhanced performance and uncovers hidden relational structures in the target data. We illustrate potential practical applications with four examples: (i) random patterns, (ii) QMNIST handwritten digits, (iii) Bayesian networks, and (iv) the pattern of stock price fluctuation pattern in S\&P500. In (i) and (ii), the strongly correlated variables were concentrated near the center of the network; in (iii), the causality pattern was identified; and in (iv), a structure corresponding to the eleven sectors emerged.
\end{abstract}

\vspace{2pc}
\section{Introduction}
Generative models thrive on the adaptability of architectures tailored to the data's characteristics. Learning the relational structure of the target data is important to obtain better generative models. The architecture of a generative model can be represented as a network that illustrates the connections between various data elements. High performance is achieved by optimizing this network structure to align with the characteristics of the data. However, the selection of the network structure is often done manually; for instance, recurrent neural networks are common for time series and sequential data. Pruning the parameters of weight matrices, as discussed in \cite{han2015learning}, is one approach for automatically selecting the network structure.

Recently, there has been extensive research on generative modeling on quantum computers. The Born machine\cite{Han:2018ey} is a generative model that utilizes the projective measurement of the quantum state. Born's rule dictates that the probability of obtaining a measurement output is the squared amplitude of the wave function of a quantum state. The theoretical quantum advantage offered by Born machines on parametrized quantum circuits was discussed in \cite{10.1038/s41534-019-0157-8, 10.1038/s41534-020-00288-9, 10.1038/s42005-024-01552-6}. Born machines are promising applications of quantum computers\cite{Liu:2018il, 10.1103/physrevapplied.16.044057}. By utilizing a tensor network representation of a wave function\cite{Orus:2014ja, Xiang_2023}, we can efficiently build Born machines even on classical computers, which has been quite successful for various test datasets\cite{Han:2018ey, Cheng:2019kq}.

A tensor network represents the partial contraction between tensors as a connection in a network. It has been utilized in various machine learning methods, such as classifiers \cite{NIPS2016_5314b967}, neural networks \cite{NIPS2015_6855456e}, and reinforcement learning \cite{PhysRevLett.132.197301}. In \cite{Han:2018ey} and \cite{Cheng:2019kq}, a one-dimensional chain (called tensor train) and a balanced tree were adopted for the tensor network representing a wave function, respectively. The  differences in the performance of resulting generative models suggest the importance of choosing the network structure. However, how we can choose the best network structure for a given data remains unclear.

In this study, we improve the scheme in \cite{Han:2018ey} and \cite{Cheng:2019kq}: The key idea is to adopt a general tree structure for the network and adaptively reconnect its branches to reflect the target data's nature better. The reconnection is based on combining and decomposing local tensors, as shown in \figref{fig:algorithm}(a). We choose a decomposition guided by the mutual information to minimize information flow. If the network structure does not fit the relational data structure, the information requires passing through unnecessarily long paths to achieve the same level of learning. In contrast, a better-fitting tree structure demands less information flow (i.e., the bond mutual information) without degrading the learning quality. Though a tree network is not the most general network structure, it strikes a balance between flexibility (e.g., the tensor train and balanced tree are special cases of the tree structure) and computational efficiency (e.g., computational time is proportional to the number of the variables). Compared to the conventional parameter adjustment approaches, such as neural network pruning, the sparsely connected network allows the proposed method to use fewer parameters during learning.

To demonstrate the virtue of the proposed method, we applied it to four data sets: (i) artificial random patterns, (ii) images of handwritten digits, (iii) causally connected random variables (Bayesian network), and (iv) the fluctuation pattern of stock prices in S\&P 500 index. The results show that the proposed generative model outperforms conventional methods when no prior knowledge of the data structure is used. In addition, the tree structure after the learning/optimization process provides insights into the (sometimes hidden) relational data structure, such as strong correlations between bits or pixels for artificial random patterns and images of handwritten digits, causal dependencies among random variables of Bayesian network, and the ``11-sector'' categories of stocks in S\&P 500 index.  The proposed method automatically learns the hidden relational target data structure and reflects it in the network structure, making it a powerful tool for data analysis even without prior knowledge of the information structure.

An optimized tensor network can also resolve significant optimization issues, the prevalence of barren plateaus in optimization landscapes, for the Born machine based on a quantum circuit\cite{10.1038/s41467-023-43908-6}. Consequently, our findings improve the potential of quantum computing in practical applications.
\section{Method}
\subsection{Optimization of tensors and network structure in a Born machine with a tensor tree}
We introduce a general tree structure in the Born machine framework in which the model distribution $p_\theta$ is defined by the Born rule,
\begin{equation}
    p_\theta(\textbf{x}) = \frac{|\langle \textbf{x}|\psi_\theta\rangle|^2}{\langle \psi_\theta|\psi_\theta\rangle},
    \label{eq:born_rule}
\end{equation}
where $|\psi_\theta\rangle$ is the quantum state of the machine. Here, we replace a real quantum system with a tensor network with a tree structure to represent $\langle \textbf{x}|\psi_\theta\rangle$. Generally, a tensor network is a partial contraction of the product of tensors. It is depicted as a graph, where the nodes (i.e., circles or ellipses) represent tensors and the lines their indices. A line connecting two nodes indicates an index to be contracted, referred to as a {\it bond}. The dimension of an index, i.e., how many possible values it may take, is called the bond dimension. We consider a tensor network with no loop, which we call {\it tensor tree}. Each component tensor has three indices. The tensor train and balanced tree used in \cite{Han:2018ey} and \cite{Cheng:2019kq}, respectively, are special cases of the general tree structure. Based on a tree structure, we can efficiently calculate various exact contractions and marginal distributions using recursive procedures.

\begin{figure*}
    \centering
    \includegraphics[width=\textwidth]{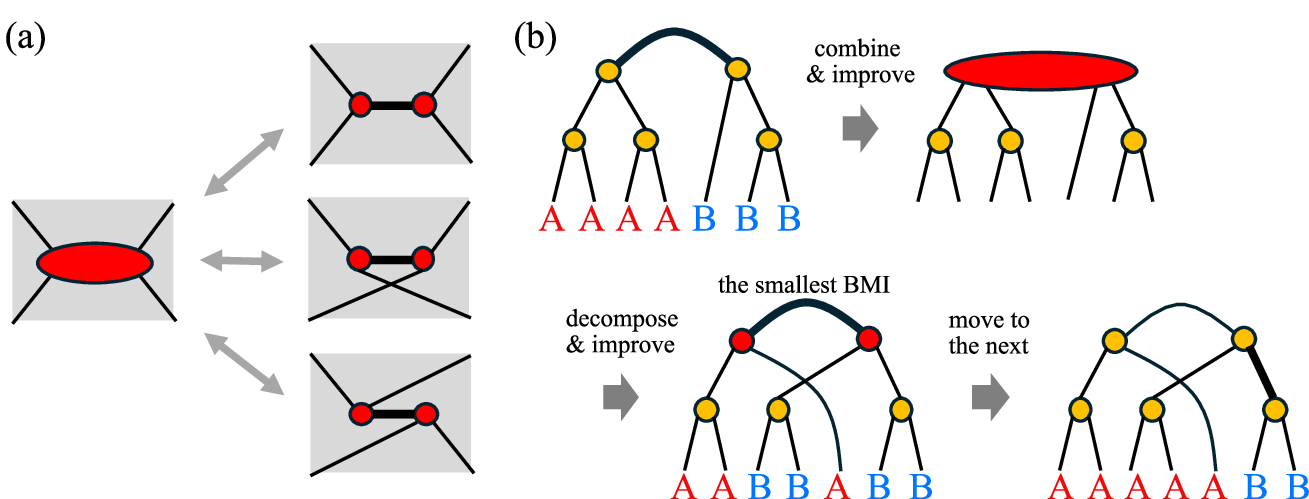}
    \caption{Procedures to be iterated for optimizing the tensors and network structure in the adaptive tensor tree (ATT) method. (a) Three candidate decompositions of a combined tensor with four legs. Before computing the bond mutual information (BMI) for each configuration, the component tensors (red) are improved.
        (b) For a target bond (the thick black line), we obtain a bigger tensor with four legs by contracting it. We choose the one with the smallest BMI among the three possible decompositions. }
    \label{fig:algorithm}
\end{figure*}

In the Born machine with a tensor tree, the mutual information that measures mutual dependence between two groups of outputs has a strict limit. By cutting a bond in a tensor tree, we decompose it into two subtrees. In addition, outputs are divided into two groups, one for each subtree. For example, for the network in \figref{fig:algorithm}(b), by cutting a thick bond, the machine's outputs are divided into parts A and B. We call the mutual information between A and B as the {\it bond mutual information} (BMI) and is given as $I(A,B)$. This quantity, $I(A,B)$ is bounded from above by the entanglement entropy of $|\psi_\theta\rangle$ for the same bipartition\cite{wu2009correlations, Convy_2022}. In addition, the entanglement entropy for the bipartition by a bond cut is less than the logarithm of the bond dimension of the cut bond\cite{Orus:2014ja, Xiang_2023}. Therefore, $I(A,B)$ is smaller than the logarithm of the bond dimension, $\chi$:
\begin{equation}
    I(A, B)=H(A) + H(B)- H(A,B) \le \ln(\chi),
    \label{eq:upper_limit_classical_mutual_information}
\end{equation}
where $H(\cdot)$ is the Shanon entropy.

These conditions provide the following guiding principle in optimizing the tree structure: when the bond dimension is fixed, avoid trees in which the BMI is close to the upper bound set by the bond dimension, which suggests putting two strongly correlated points close to each other on the tree. After combining two connected tensors in a tensor tree, we can decompose it in three ways, as shown in \figref{fig:algorithm} (a). If we choose the least BMI decomposition, we may rearrange two strongly correlated points close to each other. Therefore, we propose the branch-reconnection procedure on a tensor tree, as shown in \figref{fig:algorithm} (b). We refer to this method of adjusting the tree structure of the tensor network guided by BMI as the adaptive tensor tree (ATT) method.

\subsection{Branch-reconnection algorithm}
\label{sec:branching_reconnection}
The details of our branch-reconnection in \figref{fig:algorithm}(b) are as follows:
\begin{enumerate}
    \item Initialize the network and all tensor elements. \label{step_a}
    \item Start from a randomly selected ``virtual'' bond. (Here, a virtual bond does not directly represent an original variable.)
    \item Contract the bond and obtain a 4-leg tensor. Improve the combined tensor to approximate the empirical distribution of the data better. \label{step_b}
    \item Estimate BMI by sampling for each of the three ways of decomposition (\figref{fig:algorithm}(a)) and choose the one with the smallest BMI. Before computing BMI for each decomposition, the component tensors are improved. \label{step_c}
    \item Replace the 4-leg tensor with the chosen decomposition.
    \item If the termination condition is satisfied, terminate. \label{step_e}
    \item Move to a bond connected to one of the two new tensors and return to step (\ref{step_b}). \label{step_f}
\end{enumerate}

In steps (\ref{step_b}) and (\ref{step_c}), we improve the combined tensor and new tensors in the decomposition by employing gradient-descent updates to minimize the negative log-likelihood(NLL)\cite{Han:2018ey, Cheng:2019kq}. Assuming that the target distribution of generative modeling is an empirical data distribution, $\pi_{\textrm{emp}}$, NLL is defined as
\begin{equation}
    {\cal L} = -\frac{1}{|\cal{M}|}\sum_{\textbf{x}\in {\cal M}}\ln[p_\theta(\textbf{x})] = H(\pi_{\textrm{emp}}) + D_{\textrm{KL}}(\pi_{\textrm{emp}}||p_\theta),
    \label{eq:NLL}
\end{equation}
where ${\cal M}$ is the ensemble of data samples and $D_{\textrm{KL}}$ is a Kullback-Leibler divergence. We perform a single gradient-descent update in step (\ref{step_b}) and ten updates in step (\ref{step_c}) of the procedure in Sec.~\ref{sec:branching_reconnection}. The learning rate is set to 0.001 in this study if we do not change.

The complexity of this algorithm is $O(\chi^5)$, in which $\chi$ is the maximum bond dimension. It equals that of the two-site update of the fixed tree network method\cite{Cheng:2019kq}.

We also utilized a reconnection-based decomposition of a composite tensor to calculate the ground state of a quantum model with a tensor tree\cite{10.1103/physrevresearch.5.013031, hikihara, hikihara2}.

\subsection{Canonical form of tree tensor network}
We can define a canonical form for a tensor tree \cite{Shi:2006hz} to determine the optimal truncation from local calculations. We first choose the root edge to construct the canonical form of a given tensor tree. Then, all the tensors become isometries by recursively applying singular value decomposition (SVD) starting from the leaves. Finally, we obtain the matrix $\Lambda$ on the root edge. The canonical form of a tensor tree consists of a central weight on the root edge and isometries directed to a root.

After we optimize the composite tensor defined at a root edge to reduce an NLL for data, we can decompose it into a central weight and two isometries connected to the root edge with a restricted rank by the SVD. The decomposition by SVD is globally optimal because the other tensors are isometries in the canonical form.

Using SVD, the root's position can be moved to any neighboring edge. Our method chooses the edge to be updated as the root. After modifying the local connection and updating the tensors around the root edge, we move the root position to the neighboring edge that has not been updated for the longest time.

The Born machine is practically advantageous because there is no need to restrict the parameters to ensure the positivity of the distribution function. This is because the squared amplitude of the wave function represents it. Hence, we can use SVD to precisely decompose the whole distribution function into a tensor tree with the canonical form. Interestingly, this might contribute to the stability of our optimization process because the SVD is more stable for decomposing a tensor than the non-negative matrix factorization\cite{doi:10.1137/070709967}.

\subsection{Statistical estimation of mutual information}
From the definition of the mutual information in \eqref{eq:upper_limit_classical_mutual_information}, the mutual information for a bipartition into sub-systems $A$ and $B$ is rewritten as
\begin{equation}
    I(A,B) = \sum_{(a, b)} p(a,b) \ln \frac{p(a,b)}{p(a)p(b)},
    \label{eq:classical_mi}
\end{equation}
where $a$ and $b$ are configurations on sub-systems $A$ and $B$, respectively, $p(a,b)=p_\theta(\textbf{x}=(a,b))$ is a joint distribution, and $p(a)=\sum_b p(a,b)$ and $p(b)=\sum_a p(a,b)$ are marginal distributions. We can efficiently compute the values of $p(a)$ and $p(b)$ utilizing a tensor tree representation of a joint distribution in the canonical form. Due to the orthogonality of isometries, we can replace the isometries with identity matrices on a path from a root to open indexes of marginalized variables for constructing a marginal distribution. However, estimating the mutual information in a tensor tree state is challenging due to the need to calculate the summation for all configurations $(a, b)$. The number of configurations grows exponentially with the system size.

To estimate the mutual information approximately, we replace the average over the joint distribution with that over the empirical data distribution as
\begin{equation}
    I(A, B) \approx I_{\textrm{data}}(A, B) = \frac{1}{|{\cal M}'|}\sum_{(a, b)\in {\cal M}'} \ln \frac{p(a,b)}{p(a)p(b)},
    \label{eq:estimator_classical_mi}
\end{equation}
where ${\cal M}'$ is a set of data samples.
The estimator \eqref{eq:estimator_classical_mi} may be negative if the empirical data distribution $\pi_{\textrm{emp}}(a,b)$ differs from the joint distribution $p(a,b)$ in the early stage of optimization process. Then, we replace it with the absolute value. Such modification has no significant effect on the optimization process.

An alternative method involves estimating the average of $\ln [p(a,b)/p(a)p(b)]$ using Monte Carlo sampling. If a Born machine has a tree structure, it can efficiently calculate the marginal distribution of a random variable given the other random variable states. The conditional distribution is then used to sample random variables perfectly. Repeating these steps can generate an independent sample $(a,b)$ with probability $p(a,b)$. Consequently, we can efficiently do Monte Carlo estimation of \eqref{eq:classical_mi} as \eqref{eq:estimator_classical_mi} with ${\cal M}'$, the ensemble of samples generated from $p(a,b)$ directly.

Due to the low quality of the Born machine in its early stages, we employed the first method outlined in \eqref{eq:estimator_classical_mi} to estimate mutual information. The standard deviation of the estimation of BMI with empirical distribution is almost within $2\sim 3$\% of the average of the BMI.

In order to reduce the computational cost, we used the same set of data samples to estimate NLL and mutual information. Therefore, when the number of training data samples is large, we utilized the same mini-batch, a randomly selected subset of data samples.

\section{Results}

To demonstrate the efficiency of the ATT method, we apply it to various datasets, namely random patterns, handwritten digits images, data with probabilistic dependencies generated by Bayesian networks, and the real data on stock price fluctuation patterns in S\&P500 index.

\subsection{Random patterns and images of handwritten digits}
\begin{figure*}
    \centering
    \includegraphics[width=\textwidth]{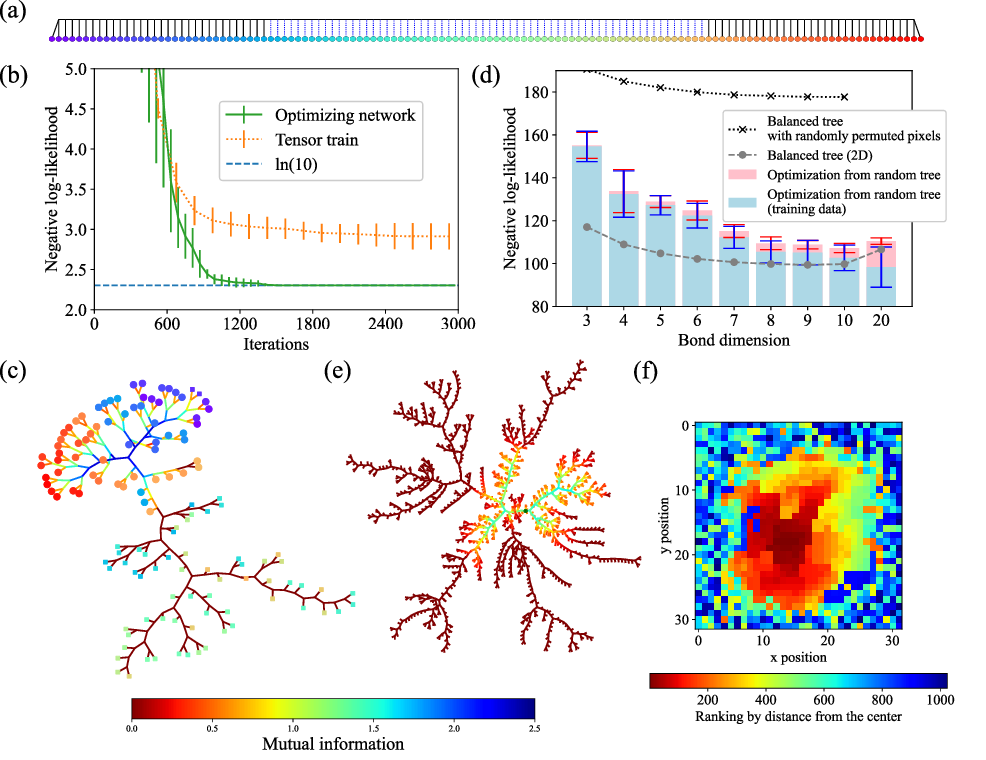}
    \caption{
        Results of random binary patterns and images of handwritten digits. (a) Tensor train with open branches corresponding to random bit variables, represented by rainbow-colored circles. Bits with dotted open branches in a middle region are fixed at 0. (b) Negative log-likelihood in learning processes for ten random binary patterns. The initial network is a tensor train. (c) Optimized network structure for ten random binary patterns. Colored circles represent random bit variables. The color of an edge indicates the amplitude of the bond mutual information. (d) Converged values of negative log-likelihood for the images of the handwritten digits vs. bond dimension. (e) Optimized network structure for the images of the handwritten digits. The color of an edge indicates the amplitude of the bond mutual information. The initial network is a random tensor tree. (f) Ranking of pixels according to their distance from the center of the tensor tree network.
    }
    \label{fig:mnist_random}
\end{figure*}

Generally, we would like a generative model to memorize finite random bit patterns accurately while effectively capturing the strong correlation among the bits. Although generative modeling with a tensor train has struggled to learn long random sequences, in \cite{Cheng:2019kq}, it was demonstrated that these issues can be resolved by a balanced tensor tree, indicating the importance of the right choice of the tensor network in addressing this problem. We consider a particularly challenging problem with the target data representing a long-range correlation to explore this issue further. Here, the data to be learned includes ten sequences, each 128 bits long. The set of the ten sequences serves as a training batch. While the left- and right-most 32 bits are generated as mutually independent random binary numbers, the bits in the intermediate part are fixed at 0 for all ten patterns, as illustrated in a tensor train in \figref{fig:mnist_random}(a). Since ten is less than the total number of possible bit patterns, it creates a strong correlation between the left- and right-most parts. In this case, we change the learning rate in the gradient-descent update in step (\ref{step_b}) and (\ref{step_c}) of the procedure in Sec.~\ref{sec:branching_reconnection} to 0.05 and set the number of iterations for termination at step (\ref{step_e}) of the procedure to 3000.

Starting with the tensor train as the initial tensor network, we compare two cases: (i) the network fixed as the initial tensor train, and (ii) the network modified by the ATT method. When the network is adaptively modified, NLL for random patterns converges to the expected lower bound, i.e., $\ln(10)$(\figref{fig:mnist_random}(b)). However, the NLL of tensor trains converges to local minimum values larger than $\ln(10)$. \figref{fig:mnist_random} (c) shows a particular realization of the final network structure. The color of the circles represents their original positions, as shown in \figref{fig:mnist_random}(a). The red and blue circles, initially far apart, are tightly clustered in the final tree, reflecting their strong correlation. The irrelevant intermediate bits (blueish green or ocher) are excluded from it and barely connected (i.e., connected by low-BMI edges).

In this example, the relevant random variables are separated in the tensor train, making the train structure disadvantageous compared to the structure derived by branch-reconnection. Below, we consider another example where the relational structure among the random variables is not immediately apparent, namely, 2D images. The balanced tensor tree outperforms the tensor train for images because its network structure aligns with the geometrical relationship among pixels \cite{Cheng:2019kq}. However, to construct a chain or balanced tree of random bits (pixels), we need to know their spatial location, i.e., they require some prior knowledge about the target data. Here, we compare the present method with fixed network methods under the condition of no prior knowledge. We apply them to images of handwritten digits from the QMNIST dataset \cite{qmnist-2019} using an initial random tensor tree or random pixel permutation, which eliminates prior knowledge of the dataset's implicit geometrical relationship.

The QMNIST dataset is an improved version of the Mixed National Institute of Standards and Technology database (MNIST) dataset\cite{deng2012mnist}, a collection of handwriting digits from 0 to 9 on $28 \times 28$ pixels. In QMNIST, the number of test images increases to the number of learning images, 60000. The intensity of a pixel ranges from 0 to 255. As in \cite{Cheng:2019kq}, by padding zero intensity pixels around an original image, we embed an original image into $32 \times 32$ pixels. We make a binary bit from a pixel's intensity. If the intensity is larger than 127, the bit is set to one; otherwise, it is set to zero. The final image consists of $32^2$ bits. For data samples ${\cal M}$ to estimate NLL and mutual information, we use a mini-batch of size 1000. However, if the mini-batch size is small, instability occurs in estimating mutual information. We set the interval for changing the mini-batch to 1000 gradient-descent updates. The number of iterations is set to $10^6$ until the pseudo-program in Sec.~\ref{sec:branching_reconnection} is terminated at step (\ref{step_e}).

We generated an initial random tree in this study as follows:
\begin{enumerate}
    \item Regard each random variable as a ``terminal tree'' consisting of only one node (leaf) with no edges. Let S be the set of all such single-node trees.
    \item Randomly take two elements from the list S and connect their roots to a new node to form a single bigger tree. Let the new node be the new tree's root, and put the new tree back into the list S.
          \label{step_random_tree}
    \item Until the list S contains only one element, repeat step (\ref{step_random_tree}).
\end{enumerate}
The final element in S is a tree with the random variables associated with its leaves.
We estimated the variance of NLLs from the five initial random trees.

\figref{fig:mnist_random}(d) displays the final values of NLL for the test handwritten digit images. Comparing the results of the fixed balanced tree with (crosses) and without (solid circles) permutation shows the significant effect of the prior knowledge about the spatial arrangement of the pixels. In contrast, network optimization with branch-reconnection (blue and red) shows performance comparable to the balanced tree for the bond dimension around 10, even with the initial random tree.
\figref{fig:mnist_random}(e) illustrates an optimized tensor tree network. The edges with strong BMI are concentrated near the center of the optimized network. We measure the distance from the center to an open edge representing a pixel to analyze the properties of the optimized network. In \figref{fig:mnist_random}(f), the color map shows pixel ranking based on the distance from the center. The compact region near the network's center is also the central region in the 2D images. Similar to the case of the separated random sequences, the generative model automatically learns the relevant relational structure among the random variables and places them close together on the tensor network.

As shown in \figref{fig:mnist_random}(b) and (d), the ATT's results show much lower NLL than the fixed tensor network models for both cases. This means that when we cannot use prior knowledge, such as the two-dimensional arrangement of the pixels, the structure optimization is efficient in reducing the NLL.

\subsection{Random variables of Bayesian networks}
A Bayesian network \cite{pearl1988probabilistic} is a stochastic model representing causal dependencies among random variables through a directed acyclic graph. A node means a random variable and a directed arrow is drawn from nodes on which a pointed random variable depends. Bayesian networks are used for modeling complex systems, such as medical diagnoses, financial forecasting, or risk analysis, where multiple variables interact in complicated ways.

However, constructing a Bayesian network from data is challenging, as this task, called structure learning \cite{strucutrelearning}, is an NP-hard problem. In a special case, if any random variable is causally dependent on only one random variable, the Bayesian network structure can be represented as a directed branching tree. In this case, Chow and Liu \cite{Chow_Liu} proposed an algorithm to construct the Bayesian network as the maximum spanning tree from the pair-wise mutual information of all possible pairs. Generally, a Bayesian network with no loop structure is called a polytree. In this case, the topology can be constructed as a maximum spanning tree \cite{Polytree}.

To the ATT method, the task of the Bayesian tree construction poses another case where random variables are mutually and causally related. While it is not clear, a priori, whether the ATT method always produces the ``correct'' tree, i.e., the Bayesian tree used in generating the data sets, we can at least expect that the correct tree is the stable solution of our method. We explain this as follows. The present algorithm decomposes the composite tensor into two pairs of smaller tensors (\figref{fig:algorithm}(a)). If two strongly related random variables of the four (with subtrees below them) are separated by this decomposition, it causes a large BMI between the two pairs. Therefore, selecting such a pairing from the three possible pairings is unlikely. In the selected pairing, the causal dependency will be stronger within each pair, not between the pairs. Hence, if we start from the topology of the polytree exactly reflecting the correct Bayesian network, the branch-reconnection procedure will not change the tree structure. Thus, the correct tree is the optimal solution for the ATT method.

We tested the ATT method using the data generated from three Bayesian polytree networks. In \figref{fig:bn_sp500h}(a-c), the Bayesian network is shown by nodes with digits and arrows. A node with a digit represents a random variable corresponding to an open index of the tensor tree. An arrow indicates the cause-to-effect direction between two random variables. The network in \figref{fig:bn_sp500h}(a) is a simple sequence with no branching. \figref{fig:bn_sp500h}(b) and (c) depict scenarios containing a branch and collision in the Bayesian network structure, respectively.
In these Bayesian networks, each random variable is binary. We generate the target data according to the causal relation. When there is a single cause, say $A$, we generate the result bit $B$ with the conditional probability $P(B=A|A)= r$. In contrast, when there are multiple causes, we first calculate the intermediate bit using the exclusive OR(XOR) of all the cause bits and generate the result bit with the same conditional probability. The present article's correlation rate $r$ is fixed to be 0.8.
For data samples, we used a mini-batch of size 1000 and updated it when every edge had been examined once with the previous mini-batch. In all cases, we use a bond dimension of 4 for a tensor tree. The optimized tensor tree is illustrated by circles without digits and bonds in \figref{fig:bn_sp500h}(a-c). Since the NLL for test data fluctuates in the optimization process when changing a mini-batch, we showed the optimized tensor tree with the lowest NLL for test data in \figref{fig:bn_sp500h}(a-c). We did $3000$ iterations until the pseudo-program in Sec.~\ref{sec:branching_reconnection} is terminated at step (\ref{step_e}). The optimized tensor tree, starting from random tensor trees, successfully captures the corresponding topology of Bayesian networks for all cases, though our results do not detect the direction of dependencies explicitly.

We note that the ATT method finds a many-body correlation among variables ``7," ``15," and ``16" in \figref{fig:bn_sp500h}(c) because the variable ``16" stochastically depends on the XOR of variables ``7" and ``15." Because the XOR cannot be represented as a two-point correlation, most conventional methods based only on the two-point correlations would fail to obtain the correct topology for \figref{fig:bn_sp500h}(c).

In the context of decoherence in a quantum circuit, the structure of a Bayesian network for measurement results mirrors that of the quantum circuit itself\cite{10.48550/arxiv.2106.15666, 10.1007/s42484-022-00095-9}. In these examples, the ATT method effectively infers the underlying structure from the data.

\subsection{Stock price fluctuation patterns in S\&P 500 index}
To evaluate the effectiveness of the ATT method for real data with an unknown structure, we built a generative model for the fluctuation pattern of stock prices. We focused on the fluctuation pattern of stock prices listed on S\&P 500, a widely recognized stock price index. We analyzed 436 stocks that have been a part of S\&P 500 index since 2010. We convert the change rate of each stock price into a binary value: 1 if it is higher than the average for all stocks and 0 otherwise. Thus, our whole dataset comprises 3589 samples, each represented by a 436-dimensional vector with binary components. To evaluate the generalization ability of the resulting generative models, we randomly select half of the sample as a training batch and the other half to evaluate the generalization ability by measuring the NLL.

\begin{figure*}
    \centering
    \includegraphics[width=\textwidth]{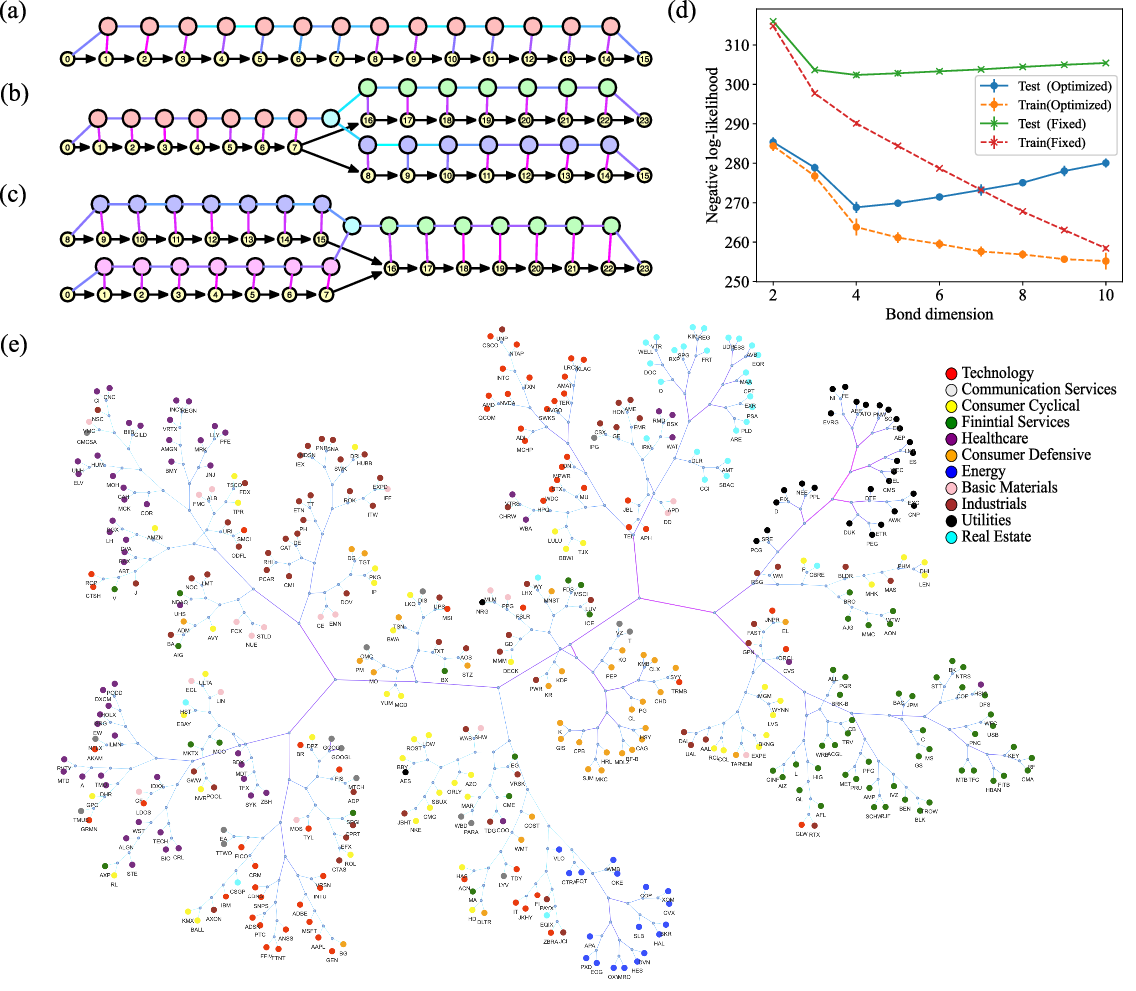}
    \caption{(a-c) Target Bayesian networks of random binary variables and corresponding tensor network structures: (a) single dependency with no branching, (b) single dependency with branching, and (c) multiple dependencies. In each diagram, the lower layer is the target Bayesian network, and the upper layer is the corresponding tensor network. All tensor trees are not schematic diagrams but actual solutions obtained by the method regardless of the initial network configurations. (d,e) The result of the adaptive tensor tree method applied to the stock price fluctuation patterns in S\&P 500 index: (d) Bond-dimension dependency of the negative log-likelihood, and (e) sample of generated tree structure at the bond dimension of 5. Companies are colored according to the sector to which they belong.}
    \label{fig:bn_sp500h}
\end{figure*}

Although the stocks are categorized into 11 sectors and even finer classifications according to the ``Global Industry Classification System,'' we do not feed the generative model with such prior knowledge. We start from a random tree and let the tree evolve using the branch-reconnection algorithm in Sec.~\ref{sec:branching_reconnection} with $2\times 10^6$ iterations until it is terminated at step (\ref{step_e}). \figref{fig:bn_sp500h}(d) represents the NLL values after learning has converged with a given bond dimension. For comparison, we also conduct training without branch-reconnections (i.e., sticking to the initial random tree throughout learning). A clear improvement achieved by the branch-reconnections can be observed in the NLL for training and testing data sets. The discrepancy between the training and testing data is the measure of the generalization performance (the smaller the better). The figure shows that the generalization performance of the generative model is at its maximum around the bond dimension 4. The bond dimension larger than five shows over-fitting.

\figref{fig:bn_sp500h}(e) shows the optimized network structure with a bond dimension of 5. The companies in S\&P 500 are classified into 11 sectors according to their business areas. The color of each circle indicates the sector to which the corresponding company belongs. The color of an edge indicates the amplitude of the BMI; purple and blue indicate strong and weak BMI, respectively. Some features of the resulting tree are independent of the initial condition. For example, companies in the same sector tend to be close and form almost single-colored sub-trees. The sector information was not used in learning, while the clustering feature is solely the result of learning. The relative location of single-colored sub-trees depends on the initial condition and is not reproducible. Regardless of the initial condition, companies in the ``utility'' sector (black) always form a sub-tree. The edge connecting the utility sub-tree to the main tree is weaker than that of edges within the sector, indicating a weak correlation with other sectors. This likely reflects the nature of the utility sector, where the demand and supply dynamics of its products, such as electricity, gas, and water, operate independently of the activities of other sectors.

\section{Discussion and Conclusion}
The proposed method has shown satisfactory performance when applied to artificial and real data sets. The optimization is done in tensor elements and the tree structure is guided by BMI. Thus, the network structure after training reflects the information flow behind the given data set; even without prior knowledge, the method successfully identified concealed information paths and adjusted the network structure to match them.
The method provides a new systematic approach to generative modeling, adapting to the information flow hidden behind the given data sets.

In the examples in the present study, we fixed the bond dimension of each edge in the tensor tree. We also defined the BMI by cutting the corresponding edge to measure the amount of information flow. The relation between the bond dimension and BMI is analogous to that between the flow capacity of a water pipe and actual flow through it. Therefore, we can adjust the bond dimension depending on BMI. For small BMI, the bond dimension can be reduced without harm. This modification could further compress the tensor tree generative models, which can be useful in future practical applications. As special cases, we may consider the Born machines using a tensor train and balanced tensor tree with automatically adjusted bond dimensions.

We let the tensor tree represent a wave function in the Born machine. Theoretically, a tensor tree state can be implemented as a quantum circuit. An optimized tensor network was utilized for preparing the parametrized quantum circuits of the Born machine to resolve serious optimization issues of the quantum circuits, the prevalence of barren plateaus in optimization landscapes\cite{10.1038/s41467-023-43908-6}. Though we have not explored this possibility in the present article, this is one of the attractive directions of future research.

In summary, the proposed method optimizes a tree structure to minimize BMI, enhancing performance and revealing hidden relational structures in the target data.
\section*{
  Data availability statement
 }
The data that support the findings of this study are openly available at the following URL/DOI: \url{https://github.com/KenjiHarada/adaptive-tensor-tree-generative-modeling}\cite{ATTCODE}.

\section*{Acknowledgments}
This work was partially supported by the joint project of Kyoto University and Toyota Motor Corporation, titled ``Advanced Mathematical Science for Mobility Society.''\cite{Mobility}
K. H. appreciates fruitful comments from T.~Hikihara, H.~Ueda, K.~Okunishi, T.~Nishino, and L.~Wang and acknowledges the support from JSPS KAKENHI (Grant No. 20K03766 and 24K06886) and a Grant-in-Aidfor Transformative Research Areas ``The Natural Laws of Extreme Universe—A New Paradigm for Spacetime and Matter from Quantum Information'' (KAKENHI Grants No. 21H05182 and No. 21H05191) from JSPS of Japan.
T. O. acknowledges the support from JSPS KAKENHI (Grant No. 23H03818 and 22K18682), the Endowed Project for Quantum Software Research and Education, The University of Tokyo (\url{https://qsw.phys.s.u-tokyo.ac.jp/}), and The Center of Innovations for Sustainable Quantum AI (JST Grant No. JPMJPF2221).
N.~K.~acknowledges the support from JSPS KAKENHI (Grant No. 23H01092).
The computation in this work has been done using the facilities of the Supercomputer Center, the Institute for Solid State Physics, the University of Tokyo.

\section*{References}
\bibliographystyle{iopart-num-long}
\bibliography{main_final}

\providecommand{\newblock}{}
\begin{thebibliography}{10}
\expandafter\ifx\csname url\endcsname\relax
  \def\url#1{{\tt #1}}\fi
\expandafter\ifx\csname urlprefix\endcsname\relax\def\urlprefix{URL }\fi
\providecommand{\eprint}[2][arXiv]{#1:\linebreak[0]#2}

\bibitem{han2015learning}
Han S, Pool J, Tran J and Dally W 2015 Learning both weights and connections for efficient neural network {\em Advances in neural information processing systems\/} vol~28

\bibitem{Han:2018ey}
Han Z~Y, Wang J, Fan H, Wang L and Zhang P 2018 {Unsupervised Generative Modeling Using Matrix Product States} {\em Physical Review\/} X {\bf 8} 031012

\bibitem{10.1038/s41534-019-0157-8}
Benedetti M, Garcia-Pintos D, Perdomo O, Leyton-Ortega V, Nam Y and Perdomo-Ortiz A 2019 {A generative modeling approach for benchmarking and training shallow quantum circuits} {\em npj Quantum Information\/} {\bf 5} 45

\bibitem{10.1038/s41534-020-00288-9}
Coyle B, Mills D, Danos V and Kashefi E 2020 {The Born supremacy: quantum advantage and training of an Ising Born machine} {\em npj Quantum Information\/} {\bf 6} 60

\bibitem{10.1038/s42005-024-01552-6}
Hibat-Allah M, Mauri M, Carrasquilla J and Perdomo-Ortiz A 2024 {A framework for demonstrating practical quantum advantage: comparing quantum against classical generative models} {\em Communications Physics\/} {\bf 7} 68

\bibitem{Liu:2018il}
Liu J~G and Wang L 2018 {Differentiable learning of quantum circuit Born machines} {\em Physical Review\/} A {\bf 98} 062324

\bibitem{10.1103/physrevapplied.16.044057}
Benedetti M, Coyle B, Fiorentini M, Lubasch M and Rosenkranz M 2021 {Variational Inference with a Quantum Computer} {\em Physical Review Applied\/} {\bf 16} 044057

\bibitem{Orus:2014ja}
Or{\'u}s R 2014 {A practical introduction to tensor networks: Matrix product states and projected entangled pair states} {\em Annals of Physics\/} {\bf 349} 117 -- 158

\bibitem{Xiang_2023}
Xiang T 2023 {\em Density Matrix and Tensor Network Renormalization\/} (Cambridge University Press)

\bibitem{Cheng:2019kq}
Cheng S, Wang L, Xiang T and Zhang P 2019 {Tree tensor networks for generative modeling} {\em Physical Review\/} B {\bf 99} 155131

\bibitem{NIPS2016_5314b967}
Stoudenmire E and Schwab D~J 2016 Supervised learning with tensor networks {\em Advances in Neural Information Processing Systems\/} vol~29

\bibitem{NIPS2015_6855456e}
Novikov A, Podoprikhin D, Osokin A and Vetrov D~P 2015 Tensorizing neural networks {\em Advances in Neural Information Processing Systems\/} vol~28

\bibitem{PhysRevLett.132.197301}
Gillman E, Rose D~C and Garrahan J~P 2024 Combining reinforcement learning and tensor networks, with an application to dynamical large deviations {\em Physical Review Letters\/} {\bf 132} 197301

\bibitem{10.1038/s41467-023-43908-6}
Rudolph M~S, Miller J, Motlagh D, Chen J, Acharya A and Perdomo-Ortiz A 2023 {Synergistic pretraining of parametrized quantum circuits via tensor networks} {\em Nature Communications\/} {\bf 14} 8367

\bibitem{wu2009correlations}
Wu S, Poulsen U~V and M\o{}lmer K 2009 Correlations in local measurements on a quantum state, and complementarity as an explanation of nonclassicality {\em Physical Review\/} A {\bf 80} 032319

\bibitem{Convy_2022}
Convy I, Huggins W, Liao H and Whaley K~B 2022 Mutual information scaling for tensor network machine learning {\em Machine Learning: Science and Technology\/} {\bf 3} 015017

\bibitem{10.1103/physrevresearch.5.013031}
Hikihara T, Ueda H, Okunishi K, Harada K and Nishino T 2023 {Automatic structural optimization of tree tensor networks} {\em Physical Review Research\/} {\bf 5} 013031

\bibitem{hikihara}
Hikihara T, Ueda H, Okunishi K, Harada K and Nishino T 2024 {Visualization of Entanglement Geometry by Structural Optimization of Tree Tensor Network} (\textit{Preprint} \eprint{2401.16000})

\bibitem{hikihara2}
Hikihara T, Ueda H, Okunishi K, Harada K and Nishino T 2025 {Improving accuracy of tree-tensor network approach by optimization of network structure} (\textit{Preprint} \eprint{2501.15514})

\bibitem{Shi:2006hz}
Shi Y~Y, Duan L~M and Vidal G 2006 {Classical simulation of quantum many-body systems with a tree tensor network} {\em Physical Review\/} A {\bf 74} 022320

\bibitem{doi:10.1137/070709967}
Vavasis S~A 2010 On the complexity of nonnegative matrix factorization {\em SIAM Journal on Optimization\/} {\bf 20} 1364--1377

\bibitem{qmnist-2019}
Yadav C and Bottou L 2019 Cold case: The lost mnist digits {\em Advances in Neural Information Processing Systems\/} vol~32

\bibitem{deng2012mnist}
Deng L 2012 The mnist database of handwritten digit images for machine learning research {\em IEEE Signal Processing Magazine\/} {\bf 29} 141--142

\bibitem{pearl1988probabilistic}
Pearl J 1988 {\em Probabilistic Reasoning in Intelligent Systems: Networks of Plausible Inference\/} (Morgan Kaufmann Publishers Inc.)

\bibitem{strucutrelearning}
Scanagatta M, Salmer^^c3^^b3n A and Stella F 2019 {A survey on Bayesian network structure learning from data} {\em Progress in Artificial Intelligence\/} {\bf 8} 425--439

\bibitem{Chow_Liu}
Chow C and Liu C 1968 {Approximating discrete probability distributions with dependence trees} {\em IEEE Transactions on Information Theory\/} {\bf 14} 462--467

\bibitem{Polytree}
Rebane G and Pearl J 1989 {The recovery of causal poly-trees from statistical data} {\em Artificial Intelligence\/} vol~3 pp 222--228

\bibitem{10.48550/arxiv.2106.15666}
Miller J, Roeder G and Bradley T~D 2021 {Probabilistic Graphical Models and Tensor Networks: A Hybrid Framework} (\textit{Preprint} \eprint{2106.15666})

\bibitem{10.1007/s42484-022-00095-9}
Liao H, Convy I, Yang Z and Whaley K~B 2023 {Decohering tensor network quantum machine learning models} {\em Quantum Machine Intelligence\/} {\bf 5} 7

\bibitem{ATTCODE}
Harada K 2024 adaptive-tensor-tree-generative-modeling \url{https://github.com/KenjiHarada/adaptive-tensor-tree-generative-modeling}

\bibitem{Mobility}
Harada K, Matsueda H and Okubo T 2024 {Application of Tensor Network Formalism for Processing Tensor Data} {\em {Advanced Mathematical Science for Mobility Society}\/} (Springer Singapore) pp 79--100

\end{thebibliography}

\end{document}